\documentclass[sigconf]{acmart}
\usepackage{xcolor}
\usepackage{soul}
\usepackage[utf8]{inputenc}
\usepackage{multirow}
\usepackage{listings}
\usepackage{pifont}
\usepackage{subcaption}

\usepackage{color}

\title[Feature Engineering and Classification Models for Partial Discharge]{Feature Engineering and Classification Models for Partial Discharge in Power Transformers}

\author{Jonathan Wang}
\affiliation{%
  \institution{Rice University}
  \city{Houston}
  \state{Texas}}
\email{jw96@rice.edu}

\author{Kesheng Wu, Alex Sim}
\affiliation{%
  \institution{Lawrence Berkeley National Laboratory}
  \city{Berkeley}
  \state{California}}
\email{kwu@lbl.gov, asim@lbl.gov}

\author{Seongwook Hwangbo}
\affiliation{%
  \institution{Hyundai Electric \& Energy Systems Co., Ltd.}
  \city{Yongin}
  \state{Korea}}
\email{smhwangbo@hyundai-electric.com}

\begin{document}

\begin{abstract}
To ensure reliability, 
power transformers are monitored for partial discharge (PD) events,
which are symptoms of transformer failure.
Since failures can have catastrophic cascading consequences, 
it is critical to preempt them as early as possible.
Our goal is to classify PDs as \textit{corona}, \textit{floating}, \textit{particle}, or \textit{void},
to gain an understanding of the failure location.

Using phase resolved PD signal data, we create a small set of features, 
which can be used to classify PDs with high accuracy.
This set of features consists of the total magnitude, the maximum magnitude, and the length of the longest empty band.
These features represent the entire signal and not just a single phase, 
so the feature set has a fixed size and is easily comprehensible.
With both Random Forest and SVM classification methods, we attain a 99\% classification accuracy, 
which is significantly higher than classification using phase based feature sets such as phase magnitude.
Furthermore, we develop a stacking ensemble to combine several classification models,
resulting in a superior model that outperforms existing methods in both accuracy and variance.

\end{abstract}

\maketitle

\section{Introduction}
\label{sec:intro}
Power transformers are a key element of electric power infrastructures.
While they have become more reliable, transformers are still susceptible to failure,
which has severe consequences for both operators and users.
To detect and prevent such failures within the large and complex power transformers,
extensive online diagnostic systems have been developed~\cite{haddad}. 
This work focuses on analyzing data produced from one such
systems to gain information about the failure.

Insulation failure is the most frequent cause of transformer failure~\cite{park}.
Weakness in the insulation system makes transformers susceptible
to external events such as lightning strikes, switching transients, and short-circuits. 
If the transformer insulation degrades to the point that it cannot withstand system events 
such as short-circuit faults or transient over-voltages~\cite{paoletti}, 
an internal arcing event known as partial discharge (PD) can occur.
As such, detecting PD events would alert transformer operators of imminent transformer failure.
Such measures could also protect other equipment connected to the transformers, 
such as Gas Insulation Switchgear (GIS) and switchboards in the substation, 
which are also expensive components of the electric power grid.

Certain types of PD are correlated with different parts of the transformer.
For example, in some transformers, corona PDs are located in the transformer bushing or insulation material.
Therefore, determining the type of PD provides a rough location for the PD source.
Machine classification is a very useful way to resolve transformer problems early on, 
and can be used by transformer operators to determine whether to halt transformer operation,
all without the need for careful examination by engineers.
By classifying the PD, we narrow down the location of the PD.
We can then install UHF sensors around the rough position
to collect PD signals to identify the precise position of the PD for repair.

In this paper, we consider four types of PDs - \textit{corona}, \textit{floating}, \textit{particle}, and \textit{void},
with the goal of analyzing PD signal data to identify what type of PD is present.
The data we examine is phase resolved, meaning it is divided into cycles of a number of phases.
Our data samples consist of 3840 points divided into 60 cycles of 64 phases.
We examine actual transformer data
and test several feature sets and classification methods
to classify PD events with high accuracy.
The contributions of our work are as follows:
\begin{itemize}
\item 
Develop a set of \textbf{meta-features}
(total magnitude, maximum magnitude, and the length of the longest empty band) 
which are more comprehensible than standard features.

\item
Test models such as Logistic Regression and Random Forest for PD classification,
and combine them to produce an \textbf{ensemble model}
that performs better than any single classification method.
\end{itemize}

\section{Related Works}
\label{sec:related}
This work focuses on classifying PDs based on voltage data.
These four PD types occur in power equipment such as transformers and GIS (Gas Insulation Switchgear).
In order to classify and analyze the characteristics of the PD types,
many experiments have been done with GIS, which has a simpler structure than transformers.

In \cite{lundgaard}, acoustic methods are proposed to detect \textit{corona} PDs in live parts of GIS.
By analyzing the effect on the particle motion and discharge characteristics in the GIS,
the discharge characteristic spectrum of linear particles with tip \textit{corona} was obtained,
and used widely for particle pattern recognition \cite{ji}.
In \cite{rostaminia}, five different types of defects 
were implemented artificially in transformer models to investigate the resulting PD signal characteristics.
However, there was a limitation as different transformers can have different failures 
depending on the transformer structure.

Unlike the above experiments, which involve manually examining signal characteristics,
machine learning based PD classification methods consist of extracting features from the data 
and training models on those features.
There are several existing feature sets such as statistical parameters or fractal features \cite{krivda}
or partial power \cite{umamaheswari}.
Some of the more common feature sets are phase magnitude, 
which consists of information regarding the magnitude of each phase of the data \cite{ma,hao}
and discrete wavelet transform \cite{ma,hunter,hao}.

Since we are working with actual transformer data as opposed to simulated data as in the above works,
we implement methods to reduce data noise.
We also present a smaller feature set of fixed size to represent the PD signal data, 
significantly simplifying the model and making the model more comprehensible.

For the classification model, the primary methods that have been tested are SVM \cite{umamaheswari,hao,hunter}
and Neural Networks \cite{krivda,hunter,ma}.
A variation of SVM, Fuzzy SVM (FSVM), was also explored in \cite{ma}.
FSVM allows for fuzzy membership in classes to resolve unclassifiable regions \cite{inoue,lin,batuwita}
by weighting the samples based on distance from the class center \cite{ma}.
In addition to these methods, we also experiment with Random Forest, Logistic Regression, and Gradient Boosting.
In addition, we combine these methods using stacking \cite{wolpert} 
to create an ensemble that utilizes the strengths of each model.

\section{Methods}
\label{sec:methods}
The goal of our work is to classify four types of PD - \textit{corona}, \textit{floating}, \textit{particle}, and \textit{void}.
We accomplish this in two steps:
\begin{itemize}
  \item Extract features to represent signal data
  \item Train machine learning model on features
\end{itemize}

\begin{figure}[tb]
\centering
\begin{subfigure}[tb]{0.48\columnwidth}
\includegraphics[width=\columnwidth]{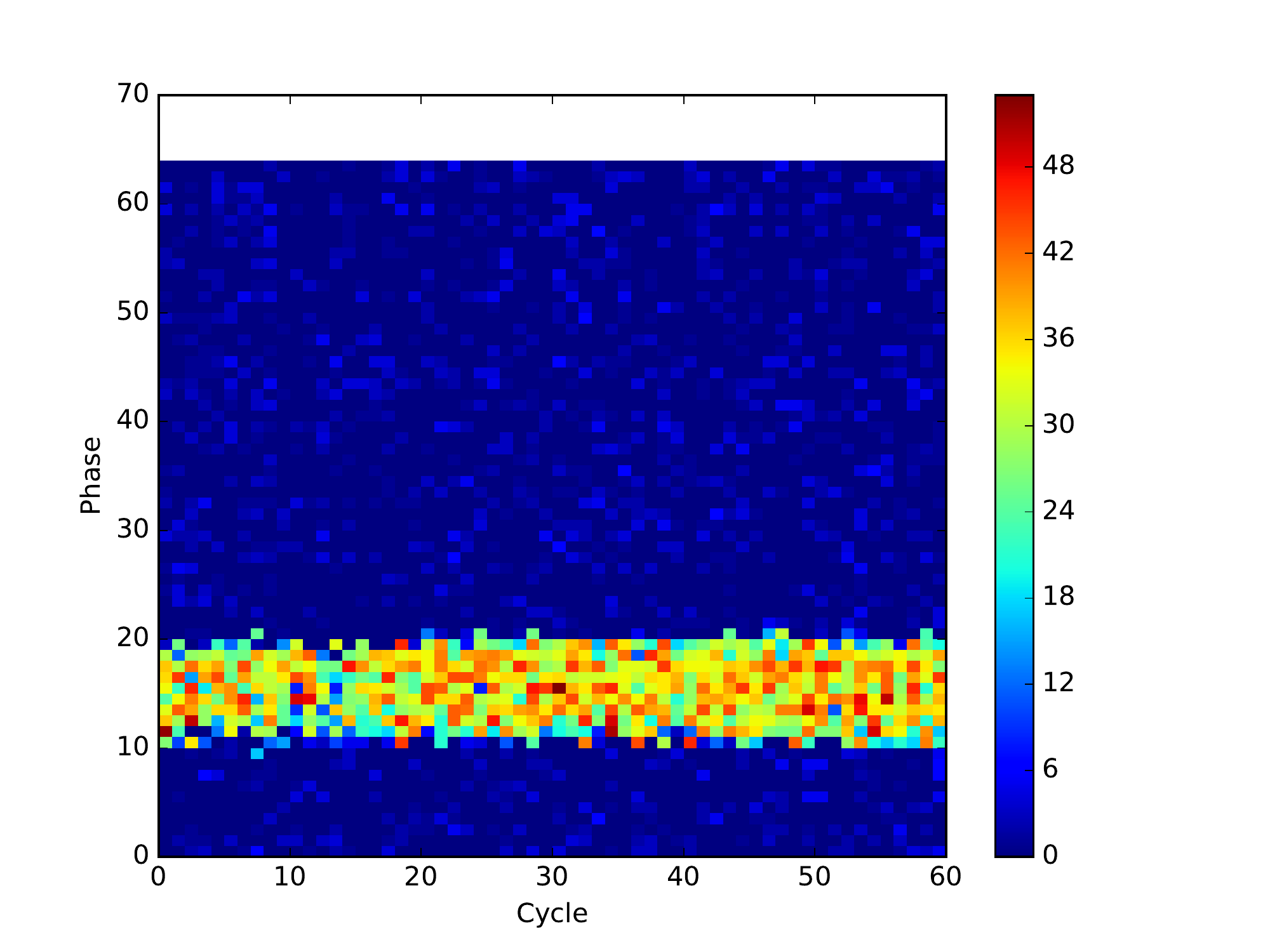}
\caption{Corona PD}
\label{fig:sample_corona}
\end{subfigure}
\hspace*{\fill}
\begin{subfigure}[tb]{0.48\columnwidth}
\includegraphics[width=\columnwidth]{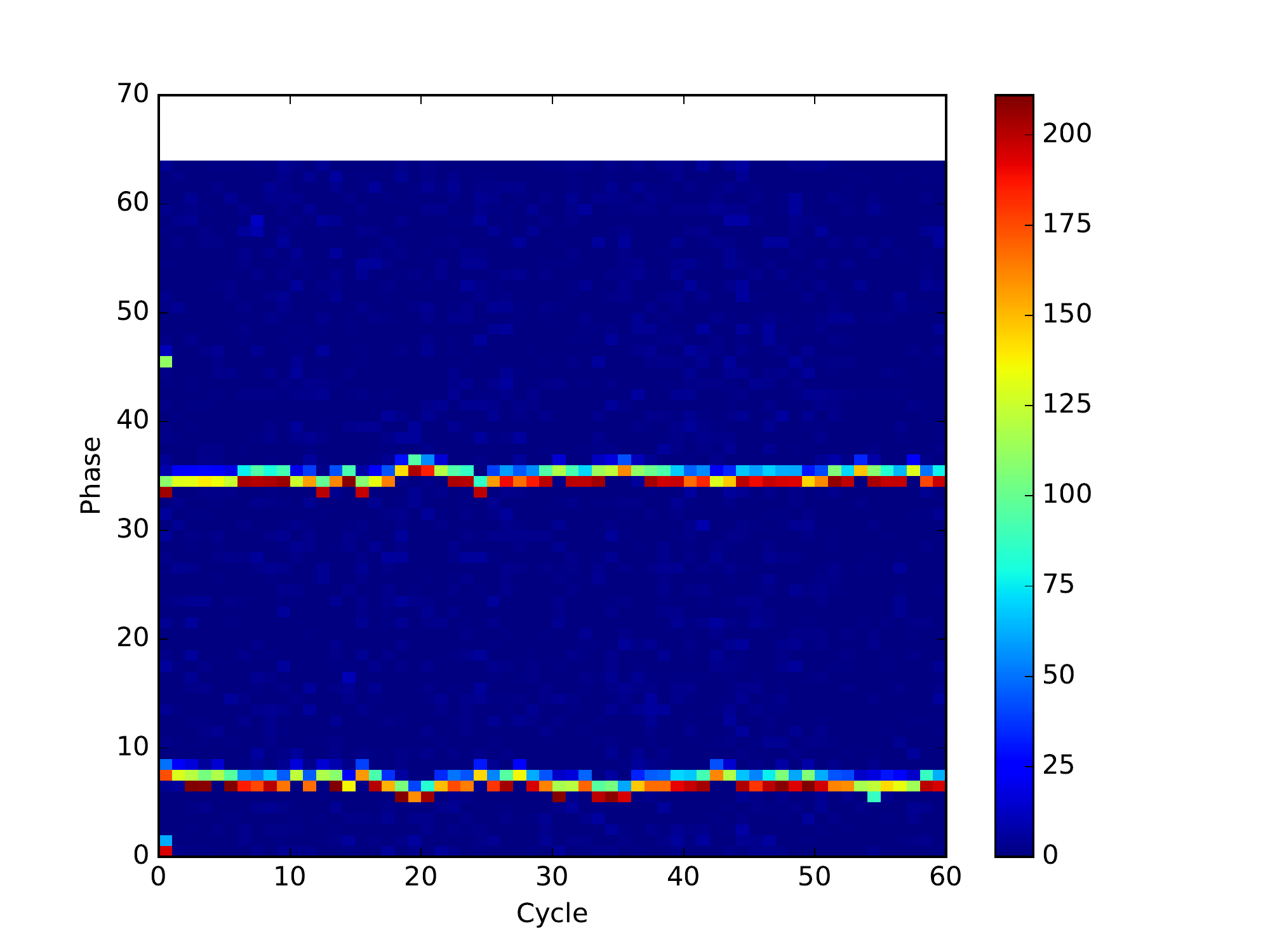}
\caption{Floating PD}
\label{fig:sample_floating}
\end{subfigure}

\begin{subfigure}[tb]{0.48\columnwidth}
\includegraphics[width=\columnwidth]{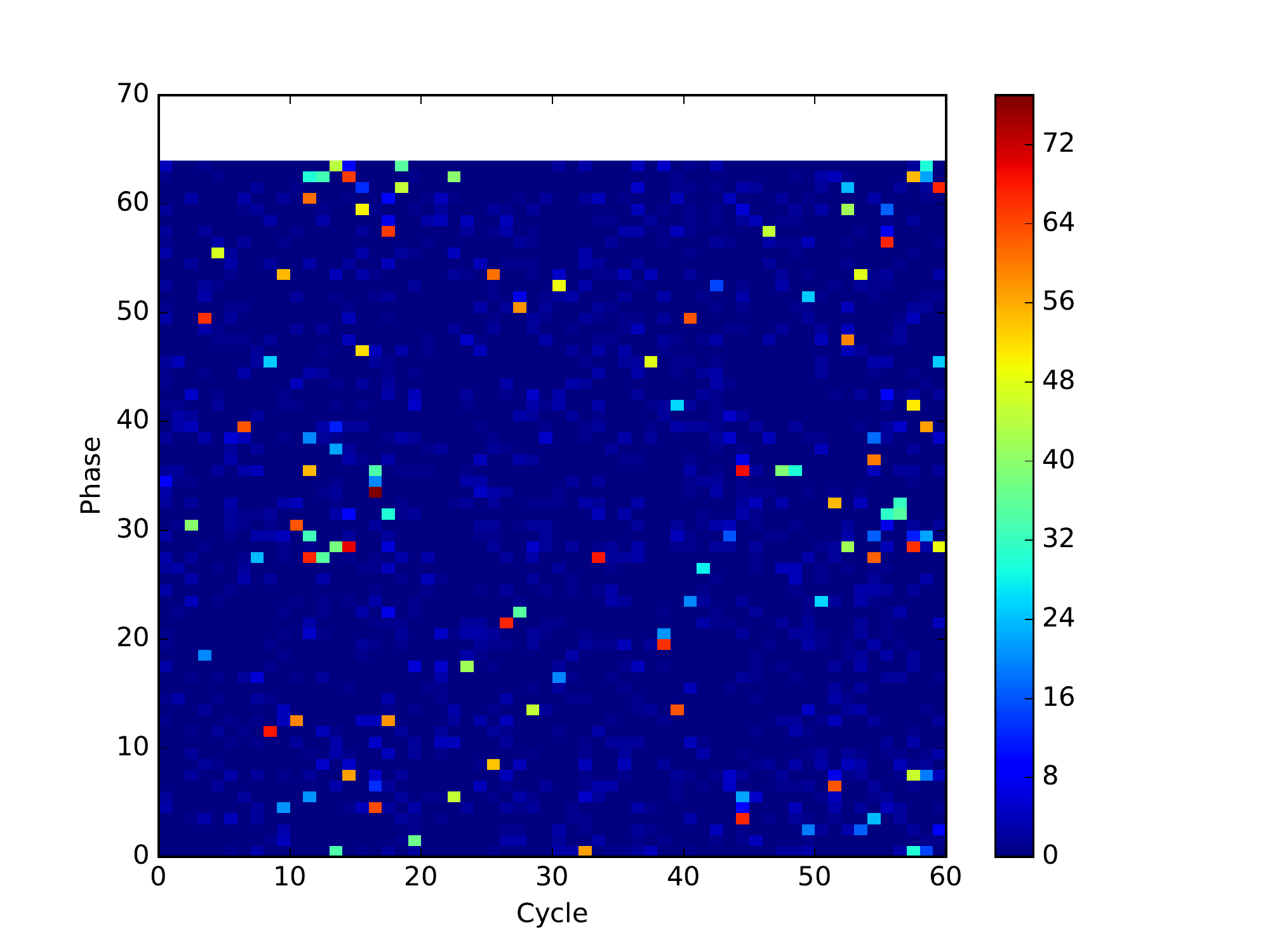}
\caption{Particle PD}
\label{fig:sample_particle}
\end{subfigure}
\hspace*{\fill}
\begin{subfigure}[tb]{0.48\columnwidth}
\includegraphics[width=\columnwidth]{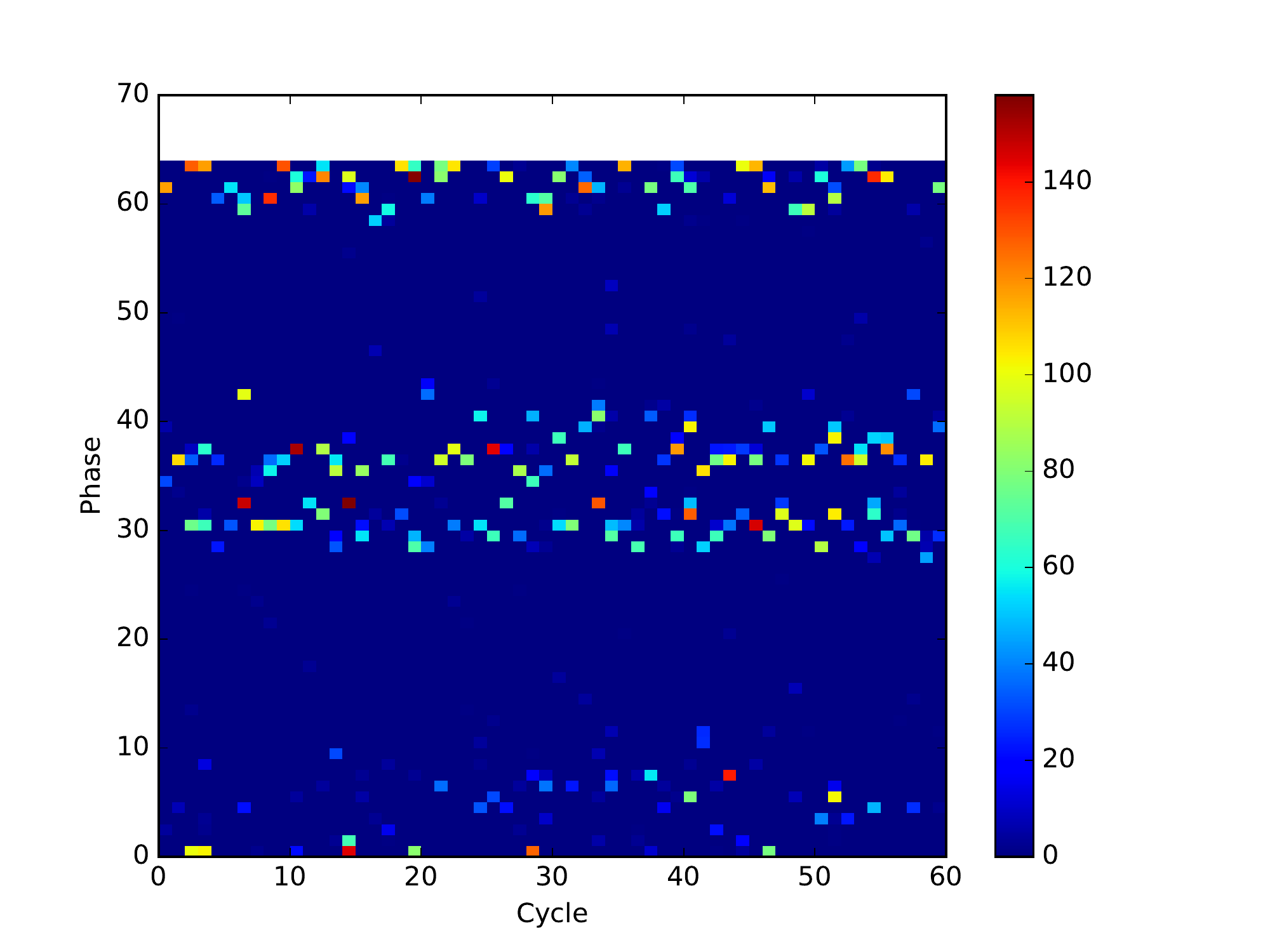}
\caption{Void PD}
\label{fig:sample_void}
\end{subfigure}
\caption{Heatmaps of Sample PDs}
\label{fig:pd_heatmaps}
\end{figure}

Our data is 328 PD signals gathered by the transformer sensors
labelled as 85 \textit{corona}, 99 \textit{floating}, 80 \textit{particle}, and 64 \textit{void}.
Each data sample contains 3840 magnitude points over one second. 
These points are broken up into 60 cycles of 64 phases.
Figure~\ref{fig:pd_heatmaps} shows heatmaps of each type of PD.
The x and y axes indicate the cycle and phase and the color indicates the magnitude at that time.

\subsection*{Feature Engineering}
\subsubsection*{Phase Magnitude}
\begin{figure}[tb]
\centering
\includegraphics[width=0.95\columnwidth]{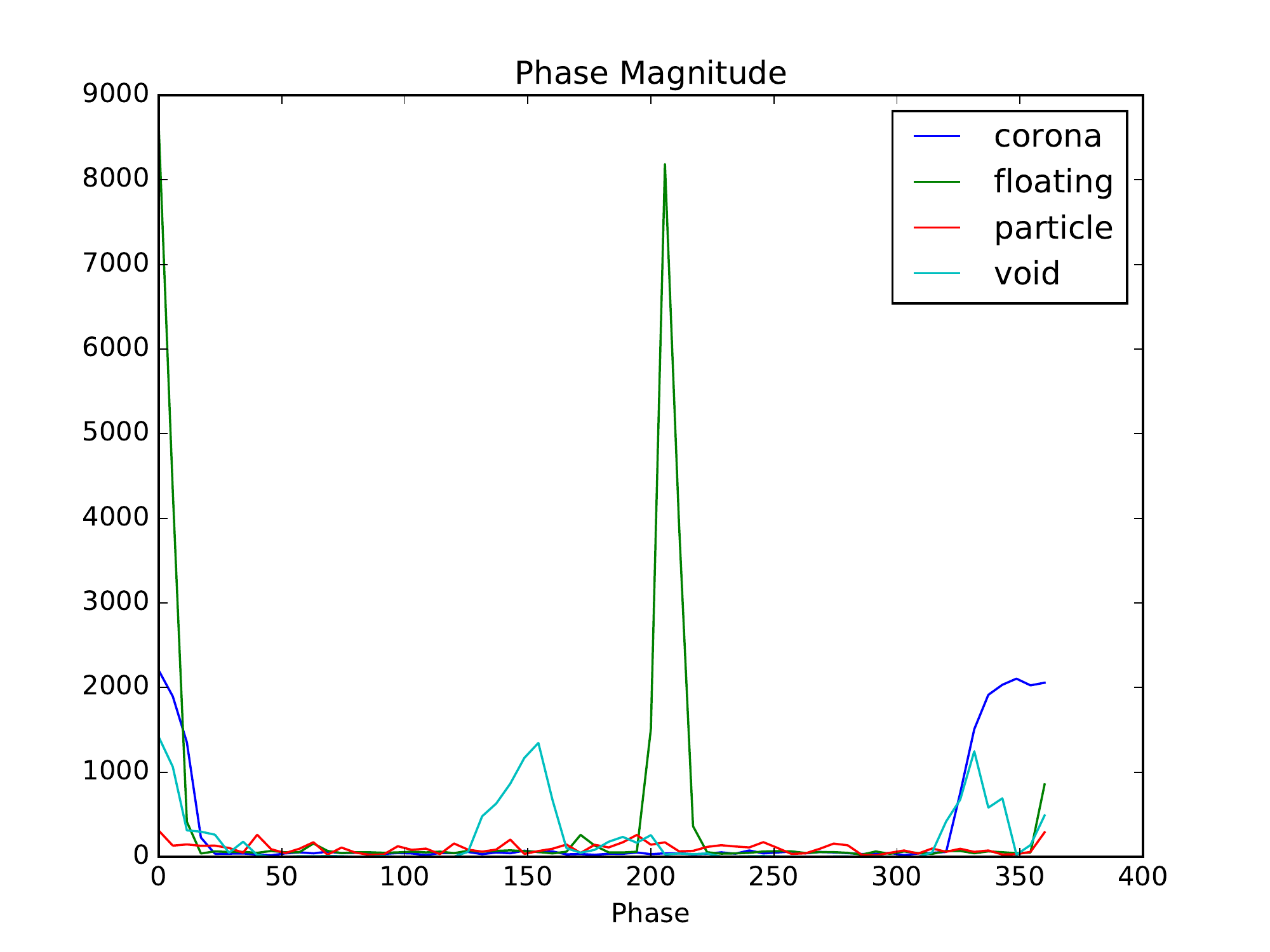}
\caption{Phase Magnitudes of Different PD Types}
\label{fig:density_comparison}
\end{figure}

We notice that there are clear patterns along the phases of each data sample.
For instance the \textit{corona} PD has a single thick band 
while the \textit{particle} PD is scattered lightly across most of the phases.
Thus, we compute the total phase magnitude for each type of PD.
The phase magnitude is the sum of the magnitudes of each phase, given by
\begin{gather*}
m_i = \sum_{j=1}^{60} M_{ij} \\
\bar{m}_p = \{m_1, ..., m_{64}\}
\end{gather*}
where $M \in \mathbb{R}^{64*60}$ is the raw magnitude data, $m_i$ is the phase magnitude for phase $i$, 
and $\bar{m}_p \in \mathbb{R}^{64}$ is the phase magnitude feature set, resulting in 64 magnitudes.
Based on the distinct phase magnitude patterns as shown in Figure~\ref{fig:density_comparison}, we can classify the PD data.

\begin{figure}[tb]
\centering
\begin{subfigure}[tb]{0.49\columnwidth}
\includegraphics[width=\columnwidth]{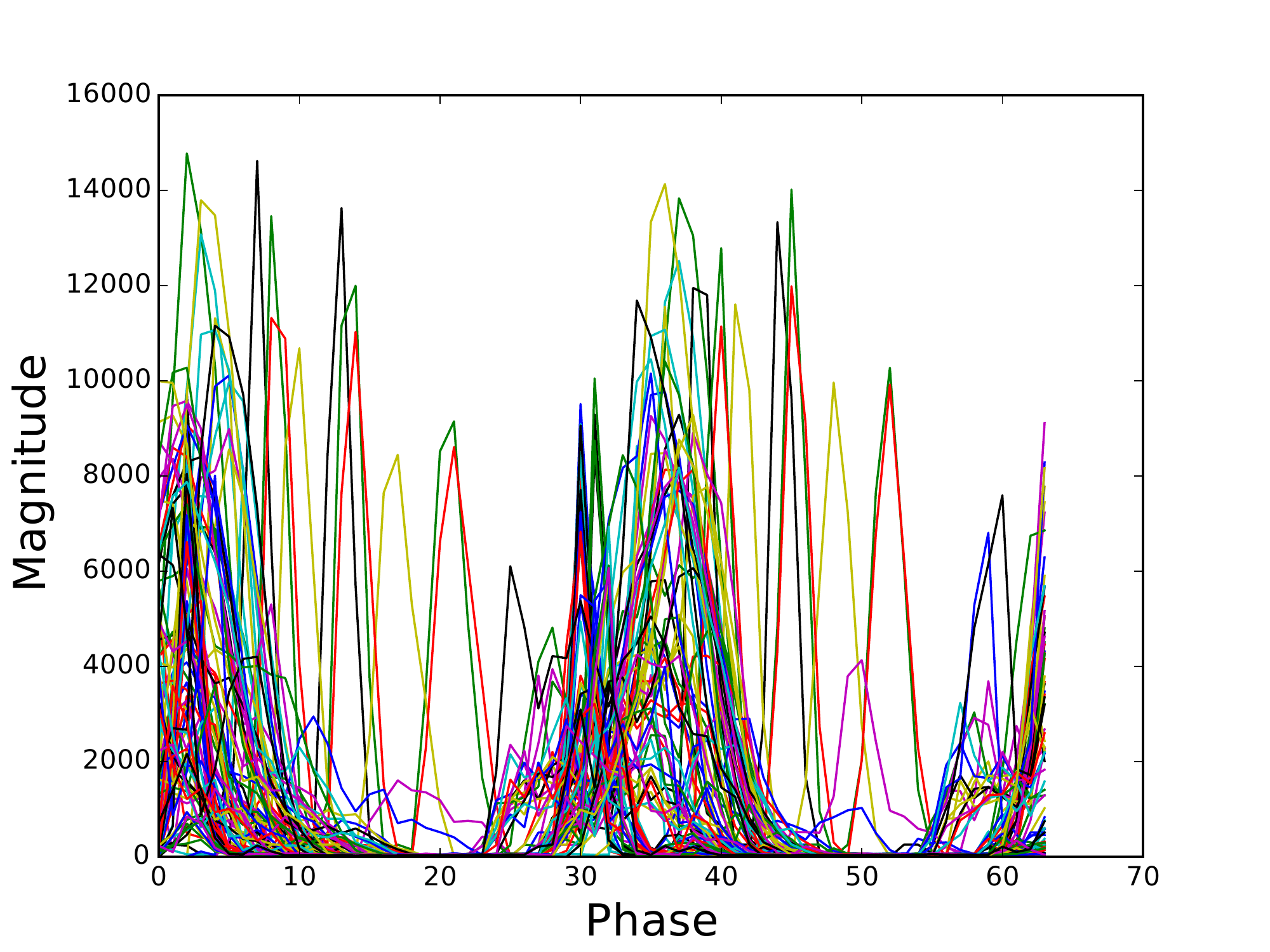}
\caption{Phases Not Aligned}
\label{fig:floating_phase_unaligned}
\end{subfigure}
\hspace*{\fill}
\begin{subfigure}[tb]{0.49\columnwidth}
\includegraphics[width=\columnwidth]{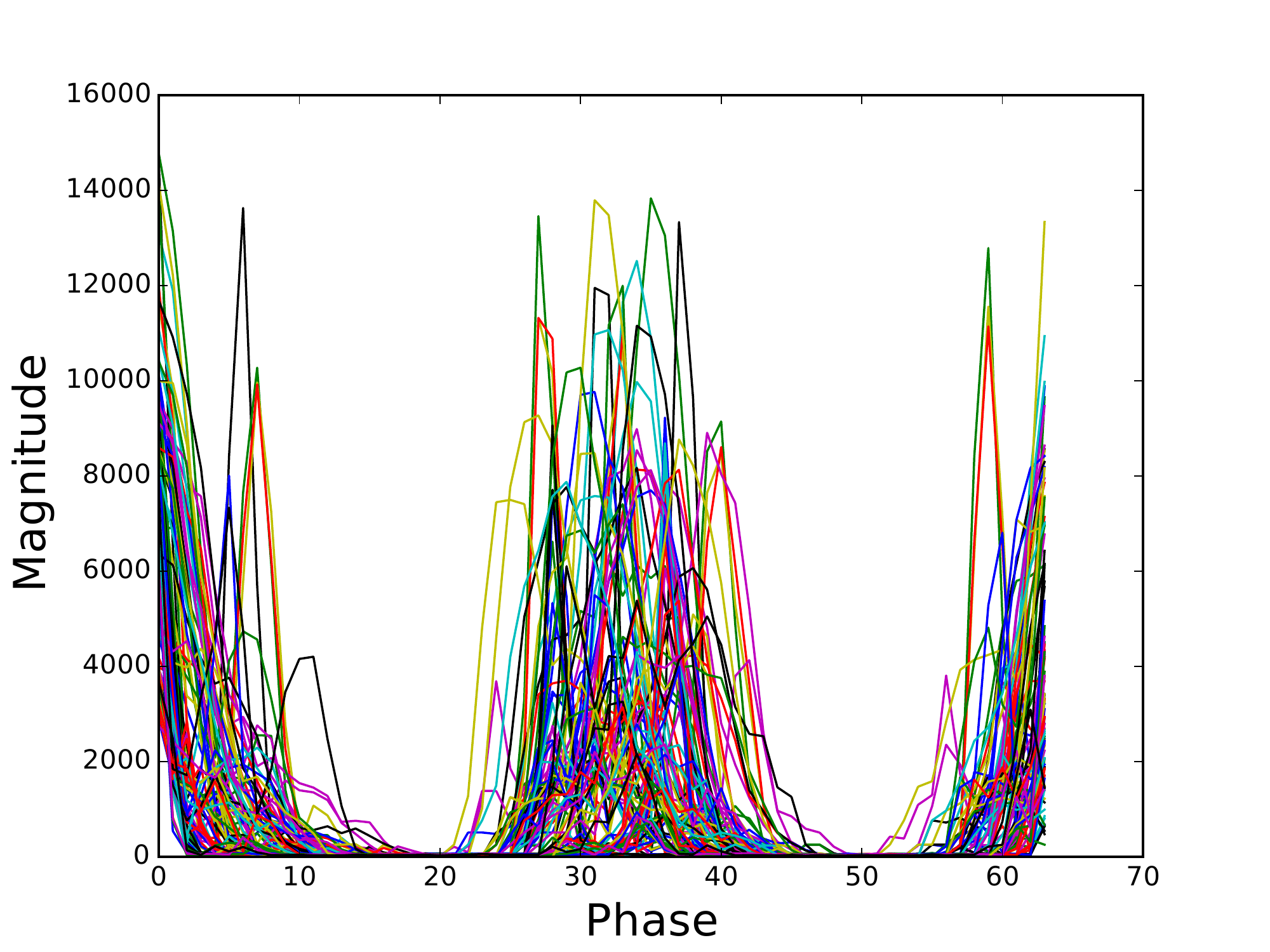}
\caption{Maximum Peak Alignment}
\label{fig:floating_phase_aligned}
\end{subfigure}
\caption{Phase Magnitude Alignment}
\label{fig:phase_alignment}
\end{figure}

There is significant misalignment in many of the samples,
where the signal does not start in the same phase as other samples.
This causes phase offsets, introducing noise to the model training.
Figure~\ref{fig:floating_phase_unaligned} illustrates the phase offset in the \textit{floating} PD data.
The outlier data samples lie outside of the regular phase magnitude peaks,
resulting in many single sample peaks across multiple phases.
To address these deviations, we implement phase alignment preprocessing 
by identifying the maximum phase magnitude and rotating that point to the start of the cycle.
While there are still a few outliers, most have been removed and the remaining effect is significantly diminished
as shown in Figure~\ref{fig:floating_phase_aligned}.

\subsubsection*{Meta-Features}
Another set of features we examine are:
\begin{itemize}
  \item the maximum magnitude out of all 3840 points
  \item the total magnitude of all 3840 points
  \item the length of the largest empty phase band
\end{itemize}
Rather than modeling the magnitude at each point or each phase,
we focus on three features that describe the overall data.
This feature set does not scale with the number of phases in the data,
and it is more comprehensible.

\begin{figure}[tb]
\centering
\begin{subfigure}[tb]{0.8\columnwidth}
\includegraphics[width=\columnwidth]{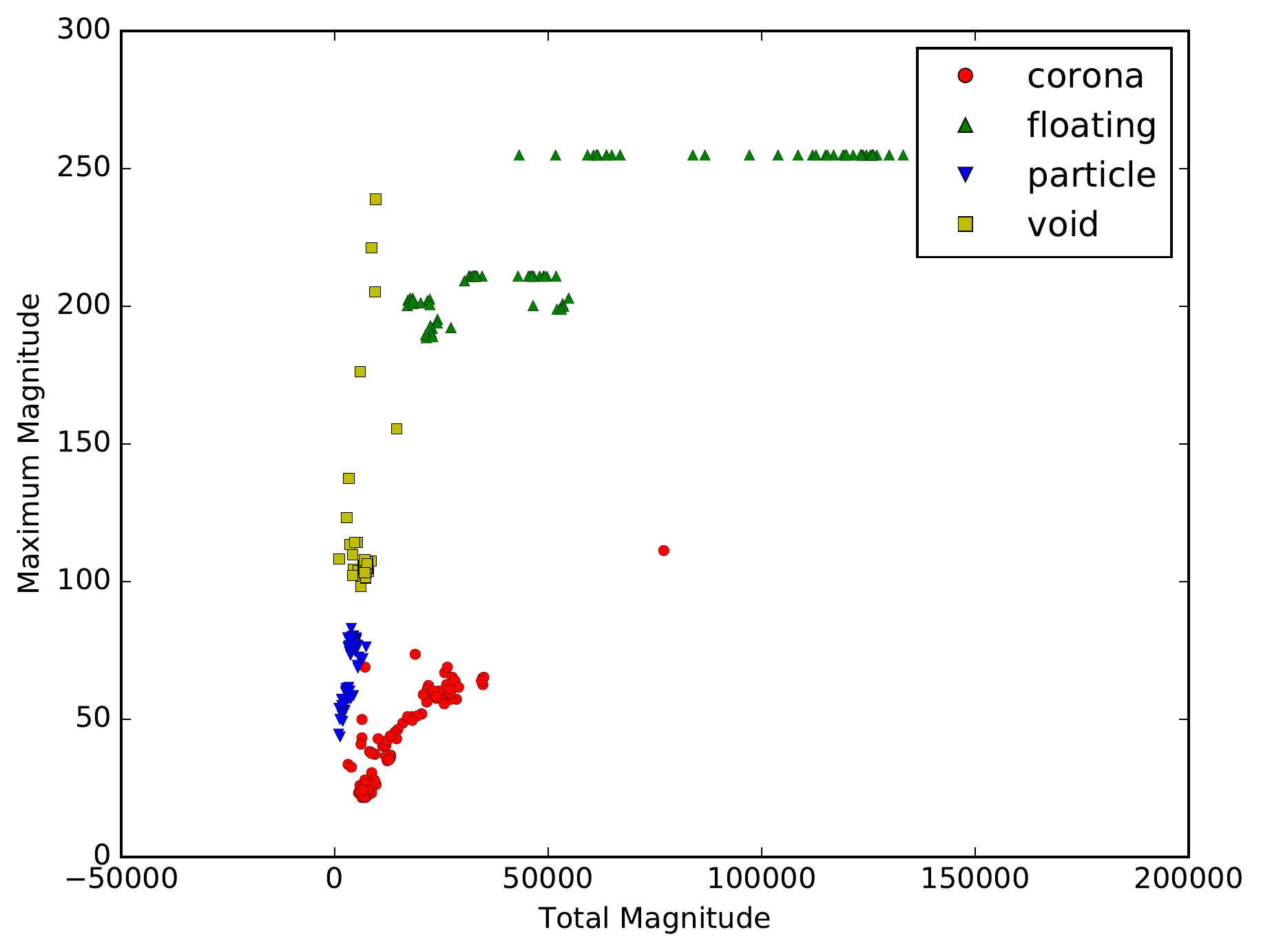}
\caption{Total and Maximum Magnitude}
\label{fig:features_total_max_mag}
\end{subfigure}
\begin{subfigure}[tb]{0.8\columnwidth}
\includegraphics[width=\columnwidth]{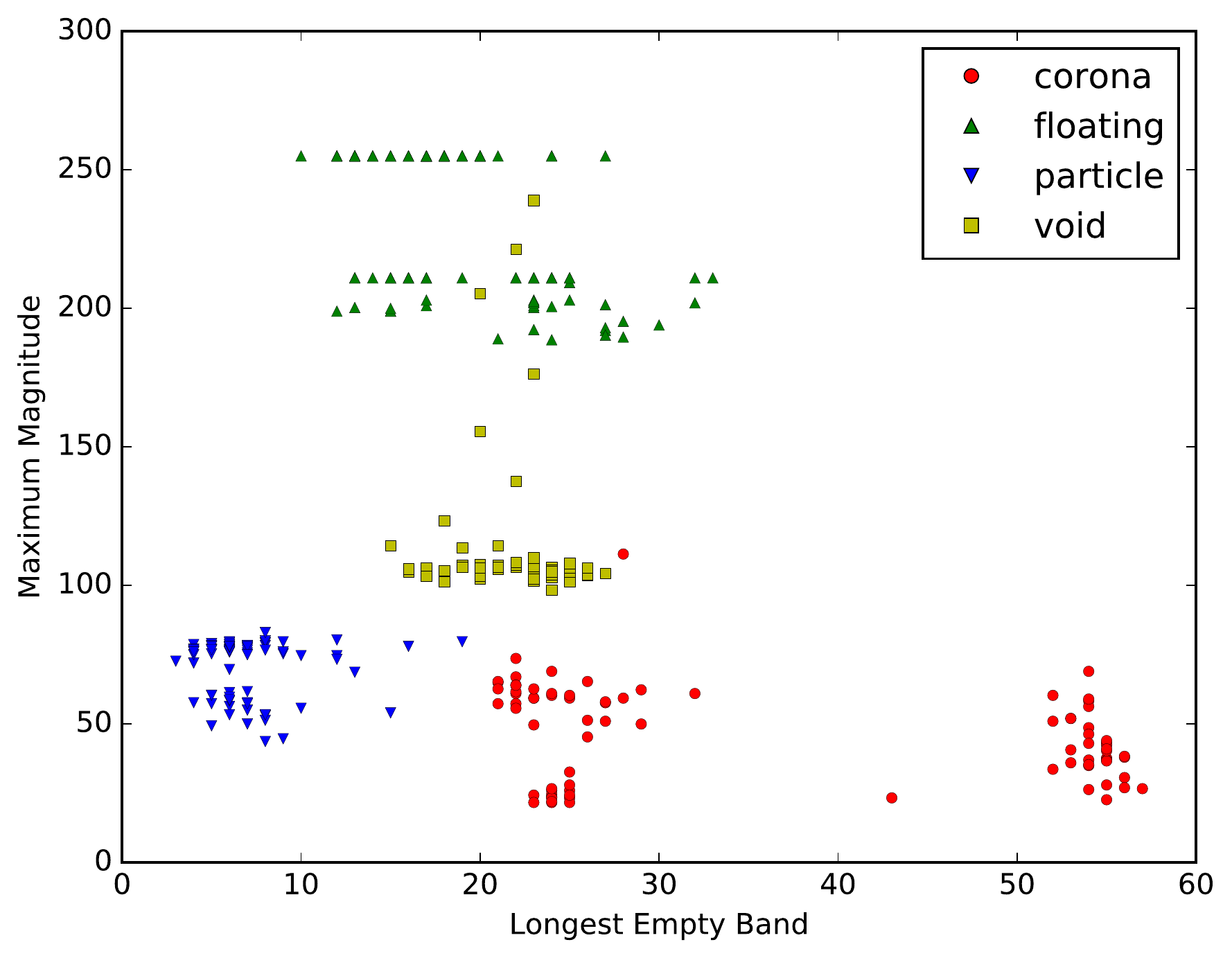}
\caption{Maximum Magnitude and Longest Empty Band}
\label{fig:features_empty_max_mag}
\end{subfigure}
\caption{PD Samples Plotted along Features}
\label{fig:meta_features}
\end{figure}

The maximum magnitude and total magnitude features are similar to features in~\cite{ma}.
However, rather than calculating the values for each phase, we consider the entire data.
We also average over the three largest magnitudes to diminish the effect of single high magnitude outliers.
Based on just these two features, we were able to classify the PDs fairly accurately,
except for the \textit{particle} PD, which had some overlap with the \textit{corona} PDs.
This observation is illustrated in Figure~\ref{fig:features_total_max_mag},
where the \textit{particle} and \textit{corona} clusters are very close and have some overlap.
To address this, we added a new feature, the length of the longest empty phase band,
which isolates the \textit{particle} PDs much better, as shown in Figure~\ref{fig:features_empty_max_mag}.
With this feature, the \textit{particle} cluster is completely separated.
In addition to these three features, we also tested several other features including total density,
although without improvement.

\begin{figure}[tb]
\centering
\includegraphics[width=\columnwidth]{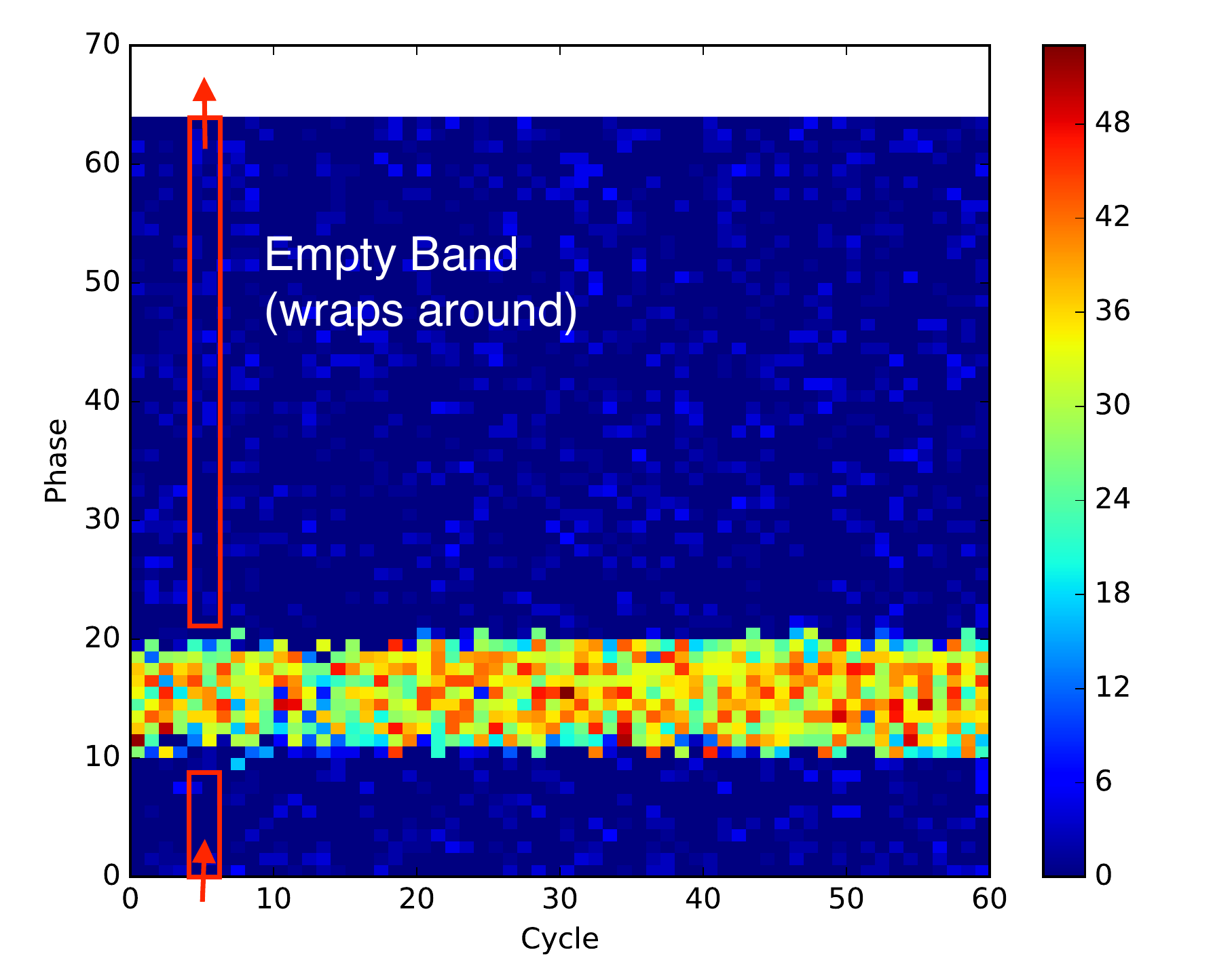}
\caption{Band of Empty Phases}
\label{fig:corona_empty}
\end{figure}

The \textbf{length of the longest empty band} quantifies the distribution of the signal.
We define an empty band as consecutive phases without significant magnitude 
as shown in Figure~\ref{fig:corona_empty}.
Significant magnitude is defined as greater than 40\% of the maximum magnitude.
This value was selected since it eliminated some outliers, 
which occurred at the original significance threshold of 50\%.
This parameter can be adjusted based on the data.

The value of this feature is the length of the longest run of empty phases,
which can be computed by
\begin{lstlisting}[language=Python, basicstyle=\small\ttfamily, columns=fullflexible, breaklines=true, frame=tb]
empty = np.sum(np.where(align_phase(data) > 0.4 * np.max(data), 1, 0), axis=1)
return max(sum(1 for _ in l) for n, l in itertools.groupby(empty))
\end{lstlisting}
This feature isolates \textit{particle} PDs, which are scattered across many phases,
resulting in very short empty phase bands.

\begin{figure}
\begin{subfigure}[tb]{0.48\columnwidth}
\includegraphics[width=\columnwidth]{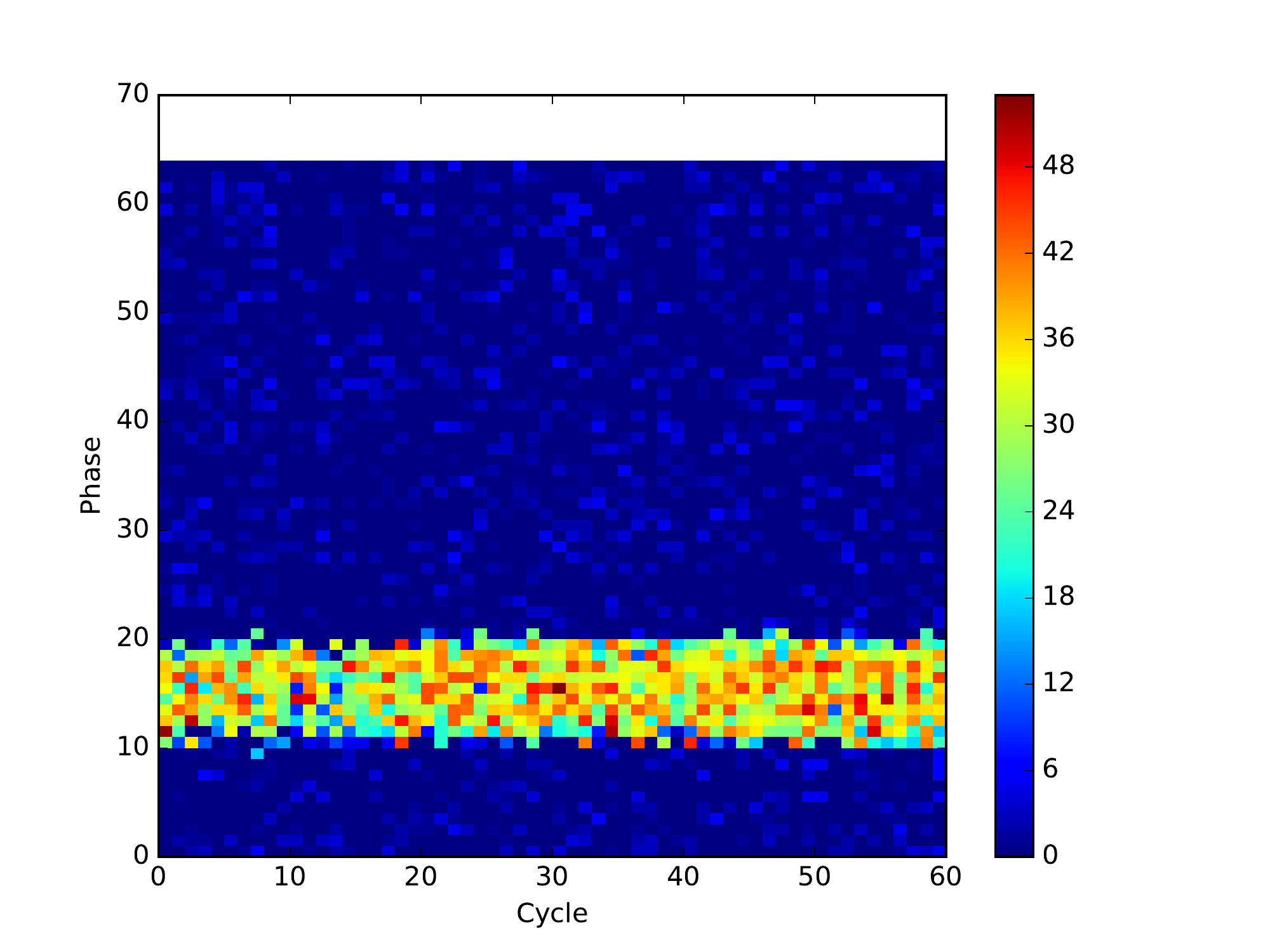}
\caption{Corona PD - Low Phase}
\label{fig:2006-06-07-15-10-28_heatmap}
\end{subfigure}
\hspace*{\fill}
\begin{subfigure}[tb]{0.48\columnwidth}
\includegraphics[width=\columnwidth]{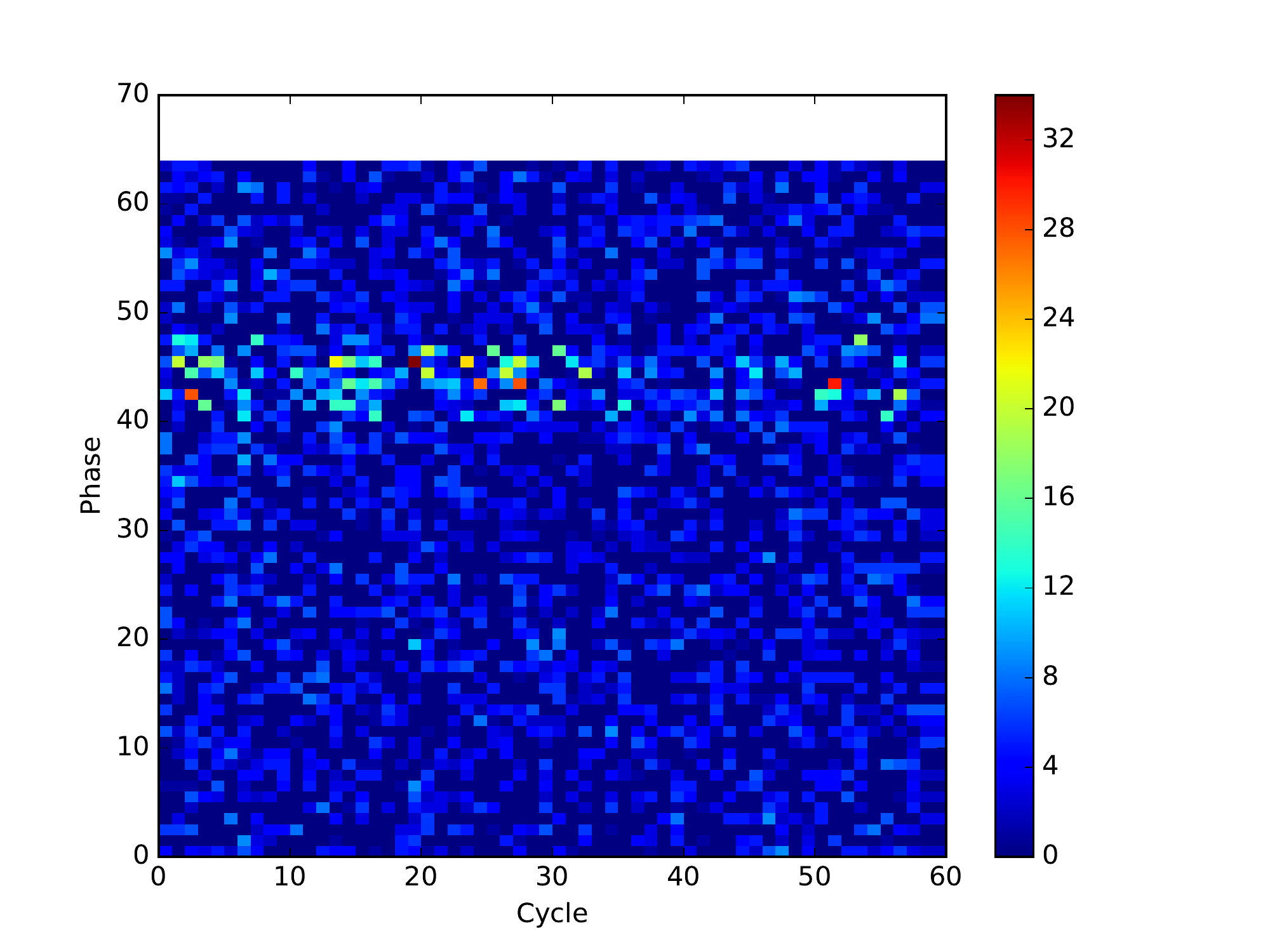}
\caption{Corona PD - High Phase}
\label{fig:2006-06-13-14-15-29_heatmap}
\end{subfigure}

\begin{subfigure}[tb]{0.48\columnwidth}
\includegraphics[width=\columnwidth]{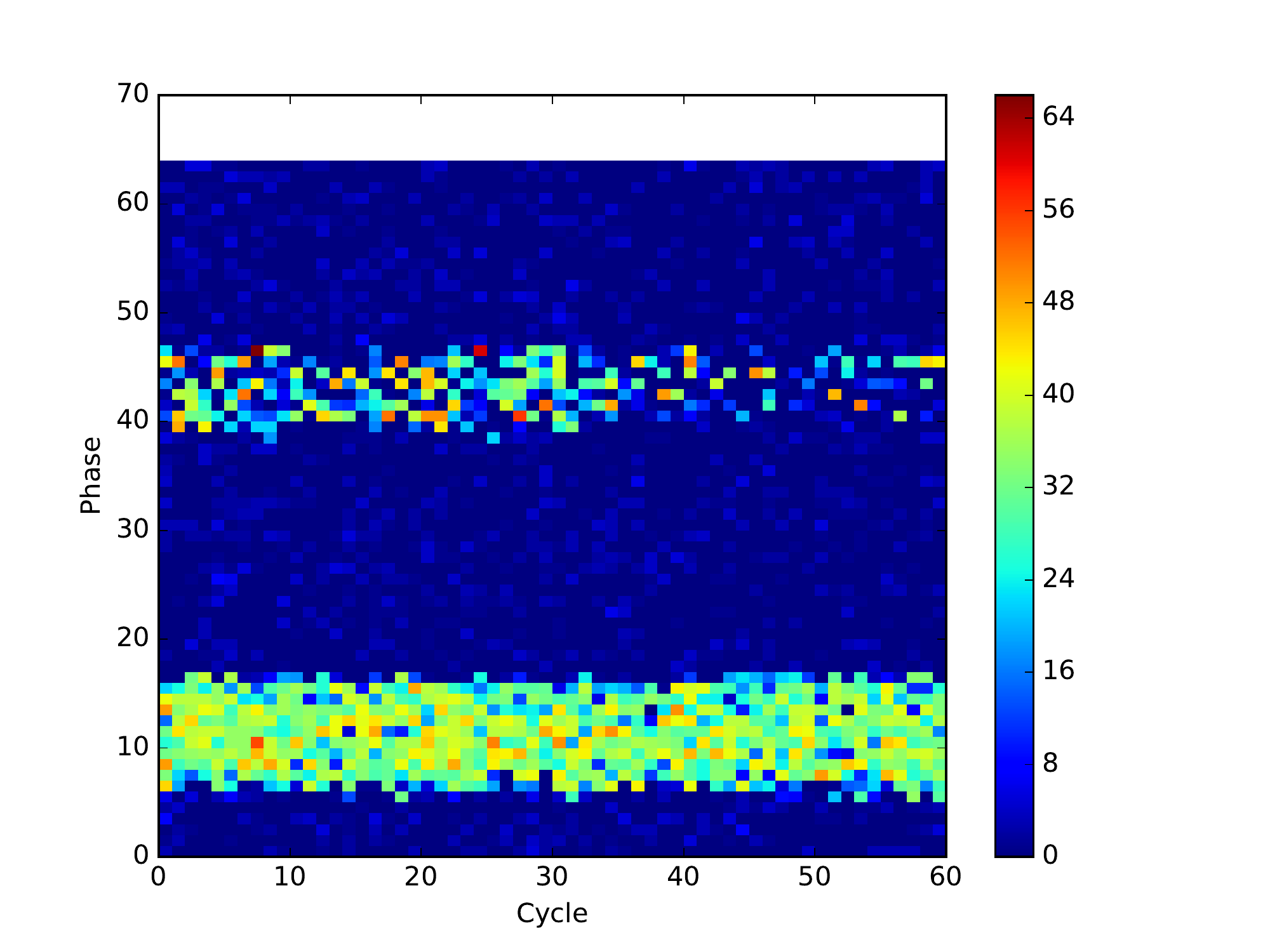}
\caption{Corona PD - Two Phase}
\label{fig:2006-06-07-15-11-06_heatmap}
\end{subfigure}
\caption{Corona PD Subtypes}
\label{fig:corona_subtypes}
\end{figure}

The meta-features are plotted with the PD type in Figure~\ref{fig:meta_features},
which shows that the data points are almost separable based on these three features.
Based on these plots, we also observe that several PD types can be sub-classified. 
For example, there are three types of \textit{corona} PDs, as shown in Figure~\ref{fig:corona_subtypes},
based on the phase at which the signal occurs.
However, since we are only looking at the four primary PD classifications,
we do not address the subclassifications in this project.

\subsection*{Prediction Models}
We test several classification methods, including Gradient Boosting, Random Forest, Logistic Regression, 
Neural Networks, SVM and FSVM.
The Fuzzy SVM is a sample-weighted SVM with the weights as an additional feature.
The method we used for sample weighting is based on the distance between the point and the cluster center similar to \cite{ma}.

For each of our experiments, we split our dataset into training and validation sets (60:40).
We train the models on the training set and score prediction accuracy based on the validation set.
For the model score, we calculate the total accuracy and the recall by PD type.
The total accuracy is the percentage of samples that have the right label.
The recall is a measure of the number of a certain PD type that are classified correctly.
This process is repeated 100 times for each model and the results are averaged to achieve more consistent results.
We also consider the standard deviation of the prediction accuracies in the 100 trials.

\subsection*{Stacking}
Certain models predict certain PD types better.
To take advantage of the strengths of each model, we implement a stacking ensemble classifier based on the meta-features.
A stacking model is comprised of two levels of classification.
We train a set of level one classifiers as in our previous experiments.
The outputs of these classifiers are then used as features for the level two classifier.
The structure of our stacking model is discussed further in Section~\ref{sec:exp}.

Stacking ensembles allow for the combination of different classifiers.
While other simpler ensembles, such as averaging or majority voting can do this as well,
these ensembles weigh each classifier equally.
In our case, Logistic Regression classifies \textit{corona} PDs very well,
so we want to value the classification of Logistic Regression more than other models for \textit{corona} classification,
but value other models more for other PD types.
Stacking handles this by training a second level classifier to weigh the importance of each level one classifier.

Several parameters are involved in designing the stacking classifier 
such as what classification methods to use in each level and what features to extract from the level one classifiers.
We consider Gradient Boosting, Random Forest, Logistic Regression, and SVM for the classifiers,
since they demonstrate reasonable classification accuracy.
While Fuzzy SVM also performed well, it requires an additional feature (sample weights),
which makes it difficult to fit into our stacking framework.
It also does not offer any advantages in classification accuracy over SVM.
Another model to consider is the classifier used to combine the results of the level one classifiers.
For this, we consider Logistic Regression, which is a standard choice, and Random Forest.

The feature parameters that we consider are 1) whether to use the probabilities or prediction from the level one classifiers
and 2) whether to include the original features in the stacking model.
The prediction is the PD type that each level one classifier determines a sample to be.
This option would result in one additional feature from each level one classifier.
The probabilities are the probabilities for a sample to be each of the four PDs determined by the classifier. 
For example, a sample could have a 70\% probability of being a \textit{corona} PD, 
a 5\% probability of being a \textit{floating} PD, a 5\% probability of being a \textit{void}, 
and a 20\% probability of being a \textit{particle} PD.
While using the prediction as a feature would result in \textit{corona} being selected,
using the four probabilities as features provides more specific information, allowing for more precise weighting.
This option requires that the level one classifiers be probabilistic classifiers, 
which is true of all of the methods we are testing.
Including the original features in the stacking model adds additional information for the stacking classifier
at the cost of diluting the results of the level one classifiers.
Using the original features also provides more information 
to help the second level classifier weight the outputs of the first level classifiers.
The second level classifier can weight classifier outputs based on the values of the original features.

\section{Experimental Results}
\label{sec:exp}
\subsection*{Phase Magnitude}
\begin{table*}[tb]
\caption{Prediction Accuracy $\pm$ Standard Deviation of Classification Models and Various Features}
\label{tab:model_accuracy}
\centering
\begin{tabular}{ | c | c | c | c | }
  \hline
  \multirow{2}{*}{Classification Method} & \multicolumn{3}{c | }{Feature Set} \\ \cline{2-4}
  & Phase Magnitude & Aligned Phase Magnitude & Meta Features \\ \hline
  Random Forest & 0.9593 $\pm$ 0.019 & 0.9699 $\pm$ 0.018 & 0.9923 $\pm$ 0.008 \\ \hline
  Gradient Boosting & 0.9475 $\pm$ 0.024 & 0.9320 $\pm$ 0.026 & 0.9807 $\pm$ 0.015 \\ \hline
  Logistic Regression & 0.9195 $\pm$ 0.021 & 0.9404 $\pm$ 0.020 & 0.9855 $\pm$ 0.012 \\ \hline
  Neural Networks & 0.6171 $\pm$ 0.213 & 0.5566 $\pm$ 0.274 & 0.2588 $\pm$ 0.047 \\ \hline
  SVM & 0.9435 $\pm$ 0.018 & 0.9673 $\pm$ 0.016 & 0.9942 $\pm$ 0.005 \\ \hline
  Fuzzy SVM (FSVM) & 0.9418 $\pm$ 0.018 & 0.9686 $\pm$ 0.016 & 0.9907 $\pm$ 0.009 \\ \hline
\end{tabular}
\end{table*}

We trained various classification models on the 328 PD samples 
using several feature sets as shown in Table \ref{tab:model_accuracy}.
Using unaligned phase magnitude, we attain a fairly high classification accuracy of about 94\%.
Random Forest performs slightly better than the next best methods - Gradient Boosting, SVM, and FSVM.
With phase alignment, the classification accuracy improves by about 3\% to 97\%.
SVM and FSVM achieve similar classification accuracies to Random Forest with this feature set.
On the other hand, Gradient Boosting performs worse with phase alignment.
In both cases, Logistic Regression is slightly behind all of the other methods,
and Neural Networks perform quite poorly, due to insufficient data samples.

With the meta-features, the prediction accuracy for Random Forest, SVM, and FSVM are all similar,
as seen in Table \ref{tab:model_accuracy}.
Gradient Boosting and Logistic Regression also attain comparable accuracies.
Again, the Neural Network has significantly lower classification accuracy than all other methods.
Compared to the phase magnitude feature sets, the meta-features have higher classification accuracy and lower variance.

\begin{table*}[!tb]
\caption{Prediction Accuracy $\pm$ Standard Deviation by PD Type}
\label{tab:type_accuracy}
\centering
\begin{tabular}{ | c | c | c | c | c | c | }
  \hline
  \multirow{2}{*}{Classification Method} & \multicolumn{4}{c | }{PD Type} & \multirow{2}{*}{Total} \\ \cline{2-5}
  & Corona & Floating & Particle & Void & \\ \hline
  SVM & 0.9915 $\pm$ 0.014 & 1 $\pm$ 0 & 0.9954 $\pm$ 0.014 & 0.9789 $\pm$ 0.042 & 0.9923 $\pm$ 0.010 \\ \hline
  Logistic Regression & \textbf{0.9997} $\pm$ 0.002 & 0.9882 $\pm$ 0.024 & 0.9680 $\pm$ 0.035 & 0.9809 $\pm$ 0.024 & 0.9847 $\pm$ 0.011 \\ \hline
  Random Forest & 0.9905 $\pm$ 0.014 & 1 $\pm$ 0 & 0.9954 $\pm$ 0.012 & 0.9832 $\pm$ 0.035 & 0.9931 $\pm$ 0.009 \\ \hline
  Gradient Boosting & 0.9672 $\pm$ 0.030 & 1 $\pm$ 0 & 0.9862 $\pm$ 0.024 & 0.9785 $\pm$ 0.035 & 0.9838 $\pm$ 0.012 \\ \hline
  Fuzzy SVM (FSVM) & 0.9859 $\pm$ 0.023 & 1 $\pm$ 0 & 0.9943 $\pm$ 0.017 & 0.9712 $\pm$ 0.029 & 0.9893 $\pm$ 0.011 \\ \hline
  Best Stacking Model & 0.9985 $\pm$ 0.007 & 1 $\pm$ 0 & 0.9984 $\pm$ 0.008 & 0.9836 $\pm$ 0.021 & 0.9961 $\pm$ 0.005 \\ \hline
\end{tabular}
\end{table*}

To better understand the model performance, we examine the classification model performance by PD type.
Table \ref{tab:type_accuracy} represents the results of this analysis.
We can see that SVM, FSVM, and Random Forest are fairly strong classifiers overall.
However, Logistic Regression performs better in classifying only \textit{corona} PDs. 
We also look at the precision of the classifications to understand the errors, 
which showed that \textit{void} PDs tend to be misclassified as \textit{floating} PDs.

\begin{figure}[tb]
\centering
\includegraphics[width=\columnwidth]{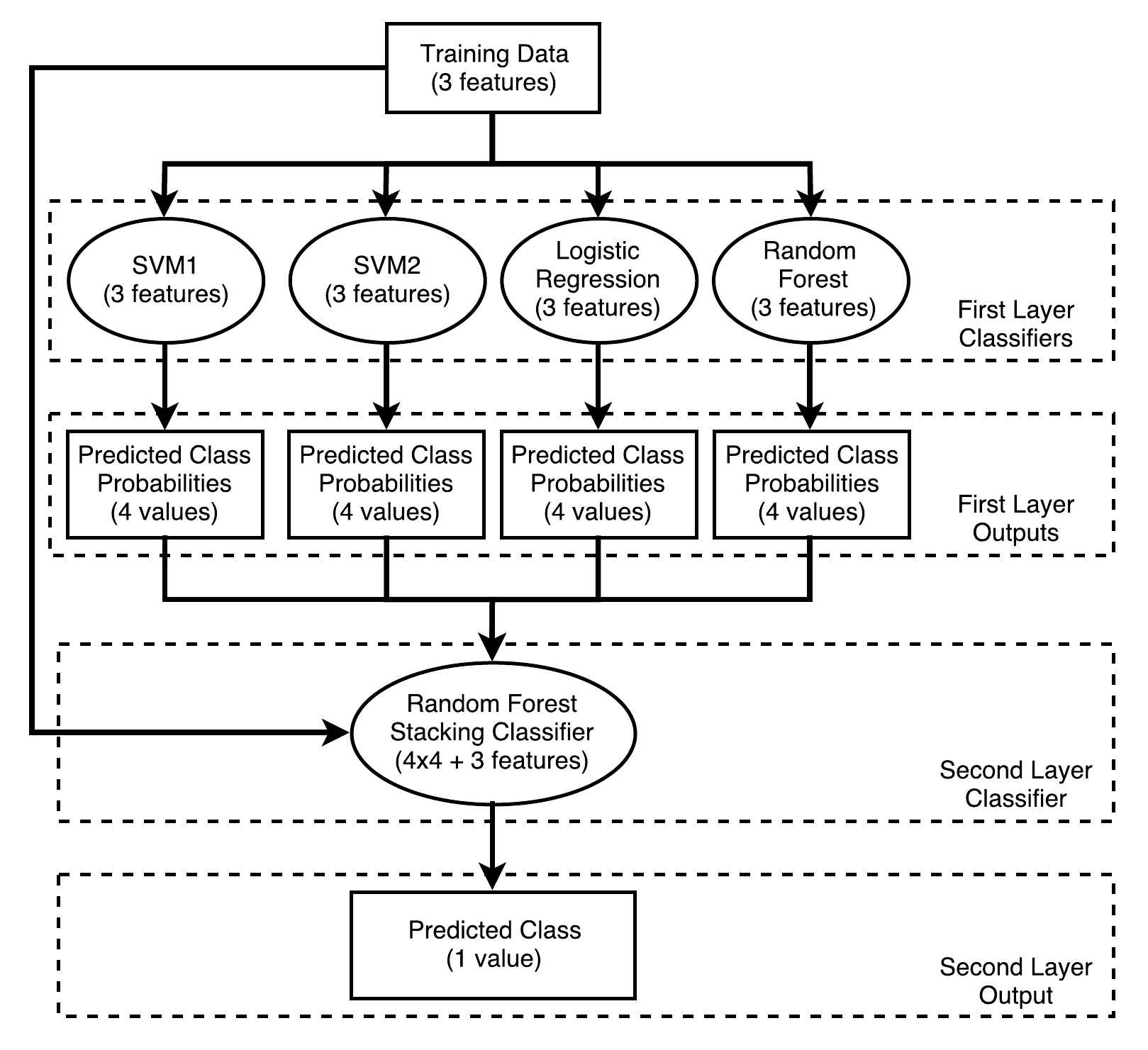}
\caption{Diagram of Final Stacking Classifier}
\label{fig:stacking_diagram}
\end{figure}

\subsection*{Stacking}
Due to the unique strengths of the various models, we combine them into an ensemble using stacking.
We try several parameter variations of the stacking classifier, but our selection is based on the observation
that Logistic Regression has much better \textit{corona} classification scores, 
and Random Forest and Gradient Boosting have more consistent \textit{floating} classification scores.
The parameters are the classifiers to stack, the model to stack with (meta-classifier), 
whether to use the model probabilities as features, and whether to include the original features in the stacked model.
The classifiers that we consider are Gradient Boosting (GB), SVM, Logistic Regression (LR), and Random Forest (RF).

We observe that Random Forest is a better meta-classifier than Logistic Regression.
Compared to the stacking model with Logistic Regression, 
using Random Forest as the meta-classifier results in higher total accuracy as well as accuracy for each PD type.
The variance for all of the accuracies are also significantly lower.

Compared to the meta-classifier features,
the selection of first level classifiers has a much larger impact on the classifier accuracy.
By examining the relative strengths of each classifier and testing various combinations,
we settle on a stacking model using 2 SVMs, Logistic Regression, and Random Forest,
which provides the best results.
This classifier is shown in Figure~\ref{fig:stacking_diagram}.
Although it does not achieve the maximum \textit{corona} accuracy of Logistic Regression, 
it offers the best total accuracy.
This model outperforms any single classification model in terms of prediction accuracy.
It also reduces the variance of the total accuracy by half and obtains the best variance for each PD type.



\section{Conclusion}
\label{sec:conclusion}
The goal of this project is to classify partial discharge (PD) events 
as \textit{corona}, \textit{floating}, \textit{particle}, or \textit{void} PDs,
to gain an understanding of the location of failure.
In order to accomplish this, we define two feature sets based on the PD data.
These feature sets are the aligned phase magnitude and the 3 meta-features - 
total magnitude, maximum magnitude, and the length of the longest empty band.

Using phase magnitude, we achieved a 94\% prediction accuracy with a variety of models.
In order to address some of the outlier samples, we use phase alignment, 
which improved prediction accuracy to 97\%.
With the meta-features, we have a more informative feature set,
and we attain an even better prediction accuracy of 99\% with most classification models.

We show that different models perform better on certain PD types.
For instance, Logistic Regression outperforms all other models in classifying \textit{corona} PDs.
We combine the strengths of each model in a stacking classifier,
resulting in an ensemble that outperforms any single model and current methods
in terms of prediction accuracy and variance.

As a result of our work, we determine a comprehensible feature set that can be used to accurately classify PD events.
We also present a stacking ensemble strategy that outperforms existing classification methods.

In future work, we plan to test our methods on additional data sets and benchmark our performance more thoroughly against various classification methods.
This data-driven solution will be integrated into an IoT-based cloud, as part of a transformer asset performance management solution,
enabling diagnosis of all transformers sold in real time to predict accidents and increase operational efficiency.
With data from a large number of various types of partial discharge patterns, 
we expect that this system can provide fast and accurate diagnosis by machine learning techniques.

\section*{Acknowledgment}
This work was supported by the Office of Science of the U.S. Department of Energy under Contract No. DE-AC02-05CH11231.

\clearpage
\bibliographystyle{ACM-Reference-Format}
\bibliography{pd_bib}

\end{document}